\documentclass[letterpaper]{article}
\usepackage{aaai}
\usepackage{times}
\usepackage{helvet}
\usepackage{courier}
\usepackage{latexsym}
\usepackage{hyperref}
\usepackage{graphicx}

\title{Learning Relational Dependency Networks for Relation Extraction}

\begin{document}
\author{Dileep Viswanathan\\
	    School of Informatics and Computing\\
  	    Indiana University\\
	    {\tt diviswan@indiana.edu}
	  \And
	Ameet Soni\\
  	Department of Computer Science\\
  	Swarthmore College\\
  {\tt soni@cs.swarthmore.edu} 
  \AND
	Jude Shavlik\\
  	Department of Computer Sciences\\
  	University of Wisconsin -- Madison\\
  	{\tt shavlik@cs.wisc.edu}
  \And
	Sriraam Natarajan\\
  	School of Informatics and Computing\\
  	Indiana University\\
  {\tt natarasr@indiana.edu}}

\nocopyright

\maketitle
\begin{abstract}
\begin{quote}

We consider the task of KBP slot filling -- extracting relation information from newswire documents for knowledge base construction.  We present our pipeline, which employs Relational Dependency Networks (RDNs) to learn linguistic patterns for relation extraction.   Additionally, we demonstrate how several components  such as weak supervision, {\tt word2vec} features, joint learning and the use of human advice, can be incorporated in this relational framework. We evaluate the different components in the benchmark KBP 2015 task and show that RDNs effectively model a diverse set of features and perform competitively with current state-of-the-art relation extraction. 
\end{quote}
\end{abstract}

\section{Introduction}

The problem of knowledge base population (KBP) -- constructing a knowledge base (KB) of facts gleaned from a large corpus of unstructured data -- 
poses several challenges for the NLP community.  Commonly, this relation extraction task is decomposed into two subtasks -- 
entity linking, in which entities are linked to already identified identities within the document or to entities in the existing KB, and slot filling, which identifies certain attributes about a target entity. 

We present our work-in-progress for KBP slot filling based on our probabilistic logic formalisms and present the different components of the system. Specifically, we employ Relational Dependency Networks~\cite{rdn}, a formalism that has been successfully used for joint learning and inference from stochastic, noisy, relational data.  We consider our RDN system against the current state-of-the-art for KBP to demonstrate the effectiveness of our probabilistic relational framework.

Additionally, we show how RDNs can effectively incorporate many popular approaches in relation extraction such as joint learning, weak supervision, {\tt word2vec} features, and human advice, among others.   We provide a comprehensive comparison of settings such as joint learning vs learning of individual relations, use of weak supervision vs gold standard labels, using expert advice vs only learning from data, etc. These questions are extremely interesting from a general machine learning perspective, but also critical to the NLP community. As we show empirically, some of the results such as human advice being useful in many relations and joint learning being beneficial in the cases where the relations are correlated among themselves are on the expected lines. However, some surprising observations include the fact that weak supervision is not as useful as expected and {\tt word2vec} features are not as predictive as the other domain-specific features. 

We first present the proposed pipeline with all the different components of the learning system. Next we present the set of $14$ relations that we learn on before presenting the experimental results. We finally discuss the results of these comparisons before concluding by presenting directions for future research. 

\section{Proposed Pipeline}
\begin{figure}
\centering
\includegraphics[width=0.45\textwidth]{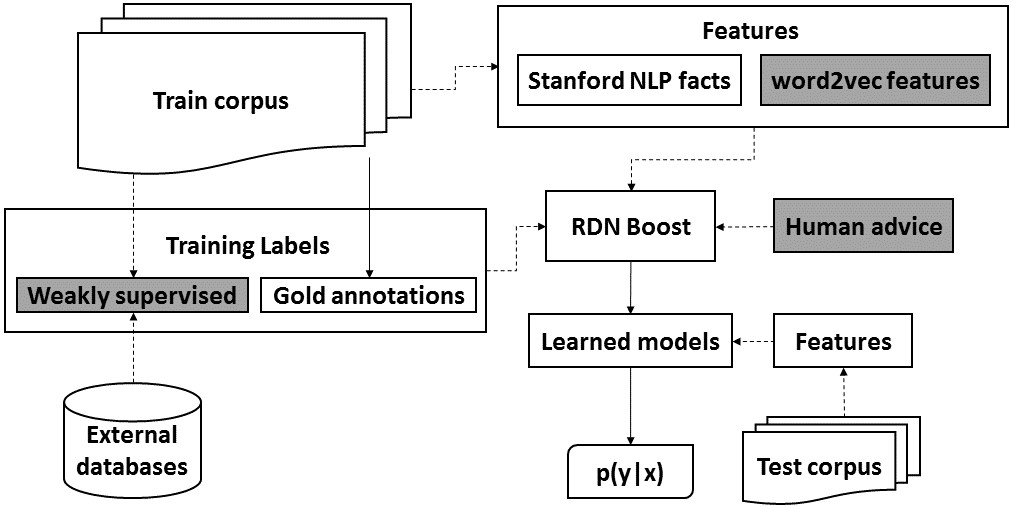}
\caption{{\bf Pipeline} Full RDN relation extraction pipeline.  Components in shaded boxes indicate proposed contributions in this work for (1) adding training labels (weak supervision), (2) enhancing feature descriptors ({\tt word2vec}), and (3) initializing the RDN with human advice rules. }
\label{fig:pipeline}
\end{figure}

We present the different aspects of our pipeline, depicted in Figure~\ref{fig:pipeline}.  We will first describe our approach to generating features and training examples from the KBP corpus, before describing the core of our framework -- the RDN Boost algorithm.  

\subsection{Feature Generation}

\begin{table}[!ht]
\centering
\small
\begin{tabular}{|l|l|}
\hline
{\bf Feature} & {\bf Description}\\\hline
wordString &  word with word id\\
wordPosition &  location of the word\\
caselessWordString & word string in lower case \\
wordLemma &  canonical form of  word\\
isNEWord & whether word is NE \\
nextWords &  two succeeding words\\
prevWords &  two preceding words\\
nextPOS &  POS for the succeeding words \\
prevPOS &   POS for the preceding words \\
nextLemmas &  canonical form of successors\\
prevLemmas &  canonical form of predecessors\\
nextNE &  succeeding NE phrases\\
prevNE &  preceding NE phrases\\
lemmaBetween & canonical form of word \\
 & occurring between two NEs \\
neBetween & word b/w two NEs is an NE\\
posBetween & POS of word b/w two NEs\\
\multicolumn{1}{|c|}{\em{Dependency Path}} & \\
rootChildLemma & canonical form of child of DPR\\
rootChildNER &  child of DPR is NE\\
rootChildPOS &  POS of child of DPR\\
rootLemma & lemma of DPR \\
rootNER &  DPR is NER \\
rootPOS & POS of DPR \\
\hline
\end{tabular}
\caption{{\bf Standard NLP Features} Features derived from the training corpus used by our learning system. POS - part of speech. NE - Named Entity. DPR - root of dependency path tree.}\label{tab:features}
\end{table}

Given a training corpus of raw text documents, our learning algorithm first converts these documents into a set of facts (i.e., features) that are encoded in first order logic (FOL).  
Raw text is processed using the Stanford CoreNLP Toolkit\footnote{\url{http://stanfordnlp.github.io/CoreNLP/}}~\cite{stanfordNLP} to extract parts-of-speech, word lemmas, etc. as well as generate parse trees, dependency graphs and named-entity recognition information. The full set of extracted features is listed in Table~\ref{tab:features}. These are then converted into features in prolog (i.e., FOL) format and are given as input to the system. 

In addition to the structured features from the output of Stanford toolkit, we also use deeper features based on {\tt word2vec}~\cite{word2vec.iclr} as input to our learning system.  Standard NLP features tend to treat words as individual objects, ignoring links between words that occur with similar meanings or, importantly, similar contexts (e.g., city-country pairs such as {\it Paris -- France} and {\it Rome -- Italy} occur in similar contexts).   {\tt word2vec} provide a continuous-space vector embedding of words that, in practice, capture many of these relationships~\cite{word2vec.iclr,word2vec.naacl}. We use word vectors from Stanford\footnote{\url{http://nlp.stanford.edu/projects/glove/}} and Google\footnote{\url{https://code.google.com/p/word2vec/}} along with a few specific words that, experts believe, are related to the relations learned. For example, we include words such as ``father'' and ``mother'' (inspired by the $parent$ relation) or ``devout'',``convert'', and ``follow'' ($religion$ relation).  We generated features from word vectors by finding words with high similarity in the embedded space.  That is, we used word vectors by considering relations of the following form: $similarWords(wordA, wordB, maxSim)$, where $maxSim$ is the  cosine similarity score between the words.  Only the top cosine similarity scores for a word are utilized.

\subsection{Weak Supervision}

\begin{table*}[t!]
\begin{center}
\begin{tabular} {|c|l|}
\hline
Weight & MLN Clause\\
\hline
$1.0$ & entityType(a, ``PER''), entityType(b, ``NUM''), nextWord(a, c), word(c, ``,''), \\
& \quad nextWord(c, b) $\rightarrow$ age(a, b)\\
$0.6$ & entityType(a, ``PER''), entityType(b, ``NUM''),
  prevLemma(b, ``age'') $\rightarrow$ age(a, b)\\
$0.8$ & entityType(a, ``PER''), entityType(b, ``PER''), nextLemma(a, ``mother'') $\rightarrow$ parents(a, b)\\
$0.8$ & entityType(a, ``PER''), entityType(b, ``PER''), nextLemma(a, ``father'') $\rightarrow$ parents(a, b)\\

\hline
\end{tabular}
\end{center}
\caption{{\bf Rules for KB Weak Supervision} A sample of knowledge-based rules for weak supervision provided by labelers.  The first value defines a weight, or confidence in the accuracy of the rule.  The target relation appears at the end of each clause. ``PER'', ``ORG'', ``NUM'' represent entities that are persons, organizations, and numbers, respectively.}
\label{CMLN}
\end{table*}

One difficulty with the KBP task is that very few documents come labeled as 
{\it gold standard labels}, and further annotation is prohibitively expensive beyond a few hundred documents.  This is problematic for discriminative learning
algorithms, like the RDN learning algorithm, which excel when given a large supervised training corpus.  To overcome this obstacle, we employ {\it weak 
supervision} -- the use of external knowledge (e.g., a database) to heuristically label examples.  Following our work in Soni et al.~\shortcite{akbc16}, we employ two approaches for generating weakly supervised examples -- distant supervision and knowledge-based weak supervision.  

Distant supervision entails the use of external knowledge (e.g., a database) to heuristically label examples.  Following standard procedure, we use three data sources -- Never Ending Language Learner (NELL)~\cite{nell}, Wikipedia Infoboxes and Freebase.  For a given target relation, we identify
 relevant database(s), where the entries in the database form {\it entity pairs} (e.g., an entry of $(Barack\ Obama, Malia\ Obama)$ for a parent database) that will serve as a seed for positive training examples.  These pairs must then be mapped to {\it mentions} in our  corpus -- that is, we must find sentences in our corpus that contain both entities together~\cite{zhang12}.  This process is done heuristically and is fraught with potential errors and noise~\cite{riedel10}. 
 
An alternative approach, knowledge-based weak supervision is based on previous work~\cite{ilp13WS,akbc16} with the following insight: labels are typically created by ``domain experts'' who annotate the labels carefully, and who typically employ some inherent rules in their mind to create examples.  For example, when identifying family relationship, we may have an \emph{inductive bias} towards believing two persons 
in a sentence with the same last name are related, or that the words ``son'' or ``daughter'' are strong indicators of a parent relation.  
We call this {\em world knowledge} as it describes the domain (or the world) of the target relation. 

To this effect, we encode the domain expert's knowledge in the form of first-order logic rules with accompanying weights to indicate the expert's confidence.  We use the probabilistic logic formalism  {\it Markov Logic Networks}~\cite{mlnbook} to perform inference on unlabeled text (e.g., the TAC KBP corpus).  Potential entity pairs from the corpus are queried to the MLN, yielding (weakly-supervised) positive examples. We choose MLNs as they permit domain experts to easily write rules while providing a probabilistic framework that can handle noise, uncertainty, and preferences while simultaneously ranking positive examples.
 
 We use the Tuffy system~\cite{tuffy} to perform inference.
The inference algorithm implemented inside Tuffy appears to be robust and scales well to millions of documents\footnote{As the structure and weights are pre-defined by
the expert, learning is not needed for our MLN}.

For the KBP task, some rules that we used are shown in Table~\ref{CMLN}.  For example, the first rule identifies any number following a person's name and separated by a comma is likely to be the person's age (e.g., ``Sharon, 42''). The third and fourth rule provide examples of rules that utilize more textual features; these rules  state  the appearance of the lemma
``mother'' or ``father'' between two persons is indicative of a parent relationship (e.g.,``Malia's father, Barack, introduced her...''). 


\subsection{Learning Relational Dependency Networks}

Previous research~\cite{SebastianRiedelJointLearning} has demonstrated that joint inferences of the relations are more effective than considering each relation individually. Consequently, we have considered a formalism that has been successfully used for joint learning and inference from stochastic, noisy, relational data called Relational Dependency Networks (RDNs)~\cite{rdn,rdnBoost}. RDNs extend dependency networks (DN)~\cite{dn} to the relational setting. The key idea in a DN is to approximate the joint distribution over a set of random variables as a product of their marginal distributions, i.e., $P(y_1,...,y_n|\mathbf{X})$ $\approx$ $\prod_i P(y_i|\mathbf{X})$. It has been shown that employing Gibbs sampling in the presence of a large amount of data allows this approximation to be particularly effective. Note that, 
one does not have to explicitly check for acyclicity making these DNs particularly easy to be learned.

In an RDN, typically, each distribution is represented by a relational probability tree (RPT)~\cite{rpt}. However, following previous work~\cite{rdnBoost}, we replace the RPT of each distribution with a set of relational regression trees~\cite{tilde} built in a sequential manner i.e., replace  a single tree with a set of gradient boosted trees. This approach has been shown to have state-of-the-art results in learning RDNs and we adapted boosting to learn for relation extraction. Since this method requires negative examples, we created negative examples by considering all possible combinations of entities that are not present in positive example set and sampled twice as many negatives as positive examples.

\begin{table}[ht]
\centering
\small
\begin{tabular}{|l|}
\hline
{\bf Advice Rules} \\\hline
Entity preceded by a number and a phrase ``year-old" \\ probably refers to age.\\ \hline
Entity present with a phrase in sentence ``who turned" \\ probably refers to age.\\ \hline
Entity1 is ``also known as'' Entity2 \\ probably refers to alternate name.\\ \hline
Entity1, ``nicknamed'' Entity2 \\ probably refers to alternate name.\\ \hline
Entity1 followed by phrase ``is a citizen of" Entity2 \\ probably refers to origin.\\ \hline
Entity followed by phrase ``is a devout" Entity2\\ probably refers to religion.\\ \hline
Entity, followed by ``a" Entity2``-based company"\\ probably refers to city/state/country of headquarters.\\ \hline
If Entity1 and Entity2 are siblings \\ then they are not parents of each other.\\ \hline
If Entity1 and Entity2 are spouses of each other \\ then they are not parents of each other.\\ \hline
\end{tabular}
\caption{{\bf Advice Rules} Sample advice rules used for relation extraction. We employed a total of 72 such rules for our 14 relations.}\label{tab:advice}
\end{table}

\subsection{Incorporating Human Advice}

While most relational learning methods restrict the human to merely annotating the data, we go beyond and request the human for advice. The intuition is that we as humans read certain patterns and use them to deduce the nature of the relation between two entities present in the text. The goal of our work is to capture such mental patterns of the humans as advice to the learning algorithm. We modified the work of Odom et al.~\shortcite{odomAIME15,odomAAAI15} to learn RDNs in the presence of advice. The key idea is to explicitly represent advice in calculating gradients.  This allows the system to trade-off between data and advice throughout the learning phase, rather than only consider advice in initial iterations.  Advice, in particular, become influential in the presence of noisy or less amout of data. 

A few sample advice rules in English (these are converted to first-order logic format and given as input to our algorithm) are presented in Table~\ref{tab:advice}. Note that some of the rules are ``soft" rules in that they are not true in many situations. Odom et al.~\shortcite{odomAAAI15} weigh the effect of the rules against the data and hence allow for partially correct rules.  


\section{Experiments and Results}

\begin{table}[t]
\begin{center}
\begin{tabular} {|l|c|c|c|}
\hline
 Relation & Gold & WS & Test \\
 \hline
 $per:age$ & 89 & 150 & 44 \\
 $per:alternateName$ & 28 & x & 18 \\
 $per:children$ & 89 & x & 23 \\
 $per:origin$ & 96 & 150 & 48 \\
 $per:otherFamily$ & 72 & 150 & 10\\
 $per:parents$ & 71 & 150  & 30 \\
 $per:religion$ & 70 & 150 & 11 \\
 $per:siblings$ & 77 & 150 & 31 \\
 $per:spouse$ & 66 & 150 & 28 \\
 $per:title$ & 158 & x & 39 \\
 $org:cityHQ$ & 69 & x & 10\\
 $org:countryHQ$ & 69 & 150 & 29\\
 $org:dateFounded$ & 70 & 150 & 17 \\
 $org:foundedBy$ & 62 & 150 & 32 \\
 \hline
\end{tabular}
\end{center}
\caption{{\bf Relations} The set of relations considered from TAC KBP.  Columns indicate the number of training examples utilized -- both human annotated (Gold) and weakly supervised (WS), when available -- from TAC KBP 2014 and number of test examples from TAC KBP 2015. 10 relations describe person entities ($per$) while the last 4 describe organizations ($org$).  }\label{numExamples}
\end{table}

We now present our experimental evaluation. We considered 14 specific relations from two categories, {\it person} and {\it organization} from the TAC KBP competition. The relations considered  are listed in the left column of Table~\ref{numExamples}. We utilize documents from KBP 2014 for training while utilizing documents from the 2015 corpus for testing.  

All results presented are obtained from 5 different runs of the train and test sets to provide more robust estimates of accuracy. We consider three standard metrics -- area under the ROC curve, F-1 score and the recall at a certain precision. We chose the precision as $0.66$ since the fraction of positive examples to negatives is 1:2 (we sub-sampled the negative examples for the different training sets).  Negative examples are re-sampled for each training run.  It must be mentioned that not all relations had the same number of hand-annotated (gold standard) examples because the $781$ documents that we annotated had different number of instances for these relations. The train/test gold-standard sizes are provided in the table, including weakly supervised examples, if available.  Lastly, to control for other factors, the default setting for our experiments is individual learning, standard features, with gold standard examples only (i.e., no weak supervision, {\tt word2vec}, advice, or advice).

Since our system had different components, we aimed to answer the following questions:
\begin{enumerate}\vspace{-0.5em}
\item[\bf Q1:] Do weakly supervised examples help construct better models?\vspace{-0.5em}
\item[\bf Q2:] Does joint learning help in some relations? \vspace{-0.5em}
\item[\bf Q3:] Are {\tt word2vec} features more predictive than standard features presented in Table~\ref{tab:features}? \vspace{-0.5em}
\item[\bf Q4:] Does advice improve performance compared to just learning from data?\vspace{-0.5em}
\item[\bf Q5:] How does our system, that includes all the components, perform against a robust baseline (Relation Factory~\cite{RothBCGK14})?
\end{enumerate}

\subsection{Weak Supervision}

To answer {\bf Q1}, we generated positive training examples using the weak supervision techniques specified earlier.  Specifically, we evaluated 10 relations as show in Table~\ref{wsresults}.  Based on experiments
from~\cite{akbc16}, we utilized our knowledge-based weak supervision approach to provide positive examples in all but two of our relations. A range of 4 to 8 rules are derived for each relation. Examples for the organization relations $countryHQ$ and $foundedBy$ were generated using standard distant supervision techniques --  Freebase databases were mapped to $foundedBy$ while Wikipedia Infoboxes provides entity pairs for  $countryHQ$. Lastly, only 150 weakly supervised examples were utilized in our experiments (all gold standard examples were utilized).  Performing larger runs is part of work in progress.

\begin{table}[t]
\begin{center}
\begin{tabular} {|l|c c|c c|}
\hline
Relation & \multicolumn{2}{|c|}{AUC ROC} & \multicolumn{2}{|c|}{F1} \\
&  G & G+WS & G & G+W \\
 \hline
$age$ & 0.90 & 0.93 & 0.64 & 0.76\\
$origin$ & 0.74 & 0.81 & 0.12 & 0.06 \\
$otherFamily$ & 0.78 & 0.83 & 0.13 & 0.23 \\
$parents$ & 0.69 & 0.62 & 0.13 & 0.25 \\
$religion$ & 0.77 & 0.73 & 0.64 & 0.54 \\
$siblings$ & 0.81 & 0.77 & 0.19 & 0.19 \\
$spouse$ & 0.83 & 0.86 & 0.31 & 0.33 \\
$countryHQ$ & 0.77 & 0.75 & 0.41 & 0.42\\
$dateFounded$ & 0.88 & 0.86 & 0.43 & 0.50 \\
$foundedBy$ & 0.85 & 0.77 & 0.50 & 0.37 \\

 \hline
\end{tabular}
\end{center}
\caption{{\bf Weak Supervision} Results comparing models trained with gold standard examples only (G) and models trained with gold standard and weakly supervised examples combined (G+WS).}\label{wsresults}
\end{table}

The results are presented in Table~\ref{wsresults}. We compared our standard pipeline (individually learned relations with only standard features) learned on gold standard examples only versus our system learned with weak and gold examples combined.  
Surprisingly, weak supervision does not seem to help learn better models for inferring relations in most cases. Only two relations -- $origin$, $otherFamily$ -- see substantial improvements in AUC ROC, while F1 shows improvements for $age$ and, $otherFamily$, and $dateFounded$.   We hypothesize that generating more examples will help (some relations produced thousands of examples), but nonetheless find the lack of improved models from even a modest number of examples a surprising result.  Alternatively, the number of gold standard examples provided may be sufficient to learn RDN models.  Thus {\bf Q1} is answered equivocally, but in the negative.

\subsection{Joint learning}
\begin{table}[t]
\begin{center}
\begin{tabular} {|l|c c|}
\hline
Relation & \multicolumn{2}{|c|}{AUC ROC} \\ 
&  IL & JL \\
 \hline
 $age$ & 0.93 & 0.93 \\
 $alternateName$ & 0.91 & 0.75 \\
 $children$& 0.75 & 0.76 \\
 $origin$ & 0.86 & 0.89\\
 $otherFamily$ & 0.88 & 0.89 \\
 $parents$ & 0.74 & 0.74 \\
 $religion$ & 0.72 & 0.79 \\
 $siblings$ & 0.79 & 0.80  \\
 $spouse$ & 0.86 & 0.87 \\
 $title$ & 0.90 & 0.89 \\
 $cityHQ$ & 0.74 & 0.73 \\
 $countryHQ$ & 0.75 & 0.79 \\
 $dateFounded$ & 0.87 & 0.86 \\
 $foundedBy$ & 0.83 & 0.86 \\
 \hline
\end{tabular}
\end{center}
\caption{{\bf Joint Learning} Results comparing models trained individually (IL) and models trained with jointly for all relations (JL).}\label{jointresults}
\end{table}

To address our next question, we assessed our pipeline when learning relations independently (i.e., individually) versus learning relations jointly within the RDN, displayed in Table~\ref{jointresults}.  Recall and F1 are omitted for conciseness -- the conclusions are the same across all metrics.  Joint learning appears to help in about half of the relations (8/14). Particularly, in person category, joint learning with gold standard outperforms their individual learning counterparts. This is due to the fact that some relations such as parents, spouse, siblings  etc. are inter-related and learning them jointly indeed improves performance. Hence {\bf Q2} can be answered affirmatively for half the relations.

\subsection{word2vec}
\begin{table}[t]
\begin{center}
\begin{tabular} {|l|c c|}
\hline
Relation & \multicolumn{2}{|c|}{AUC ROC} \\ 
&  -w2v & +w2v \\
 \hline
 $age$ & 0.93 & 0.91 \\
 $alternateName$ & 0.75 & 0.73 \\
 $children$& 0.76 & 0.79 \\
 $origin$ & 0.89 & 0.90\\
 $otherFamily$ & 0.89 & 0.78 \\
 $parents$ & 0.74 & 0.70 \\
 $religion$ & 0.79 & 0.74 \\
 $siblings$ & 0.80 & 0.75  \\
 $spouse$ & 0.87 & 0.83 \\
 $title$ & 0.89 & 0.90 \\
 $cityHQ$ & 0.73 & 0.73 \\
 $countryHQ$ & 0.79 & 0.78 \\
 $dateFounded$ & 0.86 & 0.84 \\
 $foundedBy$ & 0.86 & 0.94 \\
 \hline
\end{tabular}
\end{center}
\caption{ {\bf word2vec} Results comparing models trained without (-w2v) and with {\tt word2vec} features (+w2v).}\label{w2vresults}
\end{table}

Table~\ref{w2vresults} shows the results of experiments comparing the RDN framework with and without {\tt word2vec} features.   {\tt word2vec} appears to largely have no impact, boosting
results in just 4 relations.  We hypothesize that this 
may be due to a limitation in the depth of trees learned.  Learning more and/or deeper
trees may improve use of {\tt word2vec} features, and additional work can be done to generate 
deep features from word vectors. {\bf Q3} is answered cautiously in the negative, although future work could lead to improvements.

\subsection{Advice}
\begin{table}[t]
\begin{center}
\begin{tabular} {|l|c c| c c|}
\hline
Relation & \multicolumn{2}{|c|}{AUC ROC}  & \multicolumn{2}{|c|}{Recall} \\
&  -Adv & +Adv &  -Adv & +Adv \\
 \hline
 $age$ & 0.93 & 0.93 & 0.56 & 0.74 \\
 $alternateName$ & 0.75 & 0.77 & 0.20 & 0.16 \\
 $children$& 0.76 & 0.76 & 0.04 & 0.14 \\
 $origin$ & 0.89 & 0.88 & 0.86 & 0.82\\
 $otherFamily$ & 0.89 & 0.90 & 0 & 0.06 \\
 $parents$ & 0.74 & 0.72 & 0.15 &  0.05\\
 $religion$ & 0.79 & 0.81 & 0.51 & 0.56 \\
 $siblings$ & 0.80 & 0.81 & 0.04 & 0.00 \\
 $spouse$ & 0.87 & 0.85 & 0.06 & 0.04\\
 $title$ & 0.89 & 0.90 & 0.16 & 0.07 \\
 $cityHQ$ & 0.73 & 0.74 & 0.26 & 0.28\\
 $countryHQ$ & 0.79 & 0.77 & 0.61 & 0.62 \\
 $dateFounded$ & 0.86 & 0.86 & 0.20 & 0.05 \\
 $foundedBy$ & 0.86 & 0.84 & 0.24 & 0.25\\
 \hline
\end{tabular}
\end{center}
\caption{{\bf Advice} Results comparing models trained without (-Adv) and with advice (+Adv).}\label{advresults}
\end{table}

Table~\ref{advresults} shows the results of experiments that test the use of advice within the joint learning setting.  The use of advice improves or matches the performance of using only joint learning. The key impact of advice can be mostly seen in the improvement of recall in several relations. This clearly shows that using human advice patterns allows us to extract more relations effectively making up for noisy or less number of training examples. This is in-line with previously published machine learning literature~\cite{kbann,Fung02NIPS,KunapuliICDM13,odomAAAI15} in that humans can be more than mere labelers by providing useful advice to learning algorithms that can improve their performance. Thus {\bf Q4} can be answered affirmatively.

\subsection{RDN Boost vs Relation Factory}
\begin{table*}[t]
\begin{center}
\begin{tabular} {|l|c c| c c|c c|}
\hline
Relation & \multicolumn{2}{|c|}{AUC ROC}  & \multicolumn{2}{|c|}{Recall} & \multicolumn{2}{|c|}{F1} \\
&  RF & RDN &  RF & RDN & RF & RDN \\
 \hline
 $age$ &  0.64 & \bf 0.93 & 0.28 & \bf 0.74 & 0.44 & \bf 0.67\\
 $alternateName$ & 0.50 & \bf 0.77 & 0.00 & \bf 0.16 & 0 & \bf 0.10 \\
 $children$& 0.54 & \bf 0.76 & 0.09 & \bf 0.14 & 0.17 & \bf 0.28 \\
 $origin$ & 0.50 & \bf 0.89 & 0.00 & \bf 0.86 & 0 & \bf 0.64\\
 $otherFamily$ & 0.56 & \bf 0.90 & \bf 0.11 & 0.06 & \bf 0.24 & 0.22 \\
 $parents$ & 0.29 & \bf 0.74 & \bf 0.33 &  0.15 & \bf 0.50 & 0.31\\
 $religion$ & 0.50 & \bf 0.81 & 0 & \bf 0.56 & 0 & \bf 0.60 \\
 $siblings$    & 0.13 & \bf 0.81 & \bf 0.17 & 0.00 & 0.29 & 0.29\\
 $spouse$      & 0.57 & \bf 0.85 & \bf 0.13 & 0.04 & 0.23 & \bf 0.37\\
 $title$       & 0.67 & \bf 0.90 & \bf 0.67 & 0.07 & \bf 0.80 & 0.54\\
 $cityHQ$      & 0.38 & \bf 0.74 & \bf 0.38 & 0.28 & \bf 0.55 & 0.41\\
 $countryHQ$   & 0.57 & \bf 0.77 & 0.14 & \bf 0.62 & 0.25 & \bf 0.58\\
 $dateFounded$ & 0.67 & \bf 0.86 & \bf 0.33 & 0.05 &\bf  0.50 & 0.46\\
 $foundedBy$   & 0.20 & \bf 0.84 & \bf 0.37 & 0.25 & 0.54 & \bf 0.55\\
 \hline
\end{tabular}
\end{center}
\caption{{\bf Relation Factory vs RDN} Results comparing Relation Factory (RF) with the RDN algorithm presented in this paper.  Values in bold indicate superiour performance against the alternative approach.}\label{rfresults}
\end{table*}

Relation factory (RF)~\cite{RothBCGK14} is an efficient, open source system for performing relation extraction based on distantly supervised classifiers.  It was the top system in the TAC KBP 2013 competition~\cite{surdeanu13} and thus serves as a suitable baseline for our method.  RF is
very conservative in its responses, making it very difficult to adjust the precision levels. To be most generous to RF, we present recall for all returned results (i.e., score $> 0$). The AUC ROC, recall, and F1 scores of our system against RF are presented in Table~\ref{rfresults}.  

Our system performs comparably, and often better than the state-of-the-art Relation Factory system.  In particular, our method outperforms Relation Factory in AUC ROC across all relations. Recall provides a more mixed picture with both approaches showing some improvements -- RDN outperforms in 6 relations while Relation Factory does so in 8. Note that in the instances where RDN provides superior recall, it does so with dramatic improvements (RF often returns 0 positives in these relations).  F1 also shows RDN's superior performance, outperforming RF in most relations. Thus, the conclusion for {\bf Q5} is that our RDN framework performas comparably, if not better, across all metrics against the state-of-the-art.

\section{Conclusion}
We presented our fully relational system utilizing Relational Dependency Networks for the Knowledge Base Population task.  We demonstrated RDN's ability to effectively learn the relation extraction task, performing comparably (and often better) than the state-of-art Relation Factory system.  Furthermore, we demonstrated the ability of RDNs to incorporate various concepts in a relational framework, including {\tt word2vec}, human advice, joint learning, and weak supervision.  Some surprising results are that weak supervision and {\tt word2vec} did not significantly improve performance. However, advice is extremely useful thus validating the long-standing results inside the Artificial Intelligence community for the relation extraction task as well. Possible future directions include considering a larger number of relations, deeper features and finally, comparisons with more systems.  We believe further work on developing {\tt word2vec} features and utilizing more weak supervision examples may reveal further insights into how to effectively utilize such features in RDNs.

\bibliographystyle{aaai}
\bibliography{aaairdn}

\begin{thebibliography}{}

\bibitem[\protect\citeauthoryear{Blockeel and Raedt}{1998}]{tilde}
Blockeel, H., and Raedt, L.~D.
\newblock 1998.
\newblock Top-down induction of first-order logical decision trees.
\newblock {\em Artificial intelligence} 101(1):285--297.

\bibitem[\protect\citeauthoryear{Carlson \bgroup et al\mbox.\egroup
  }{2010}]{nell}
Carlson, A.; Betteridge, J.; Kisiel, B.; Settles, B.; Jr., E.~H.; and
  T.Mitchell.
\newblock 2010.
\newblock Toward an architecture for never-ending language learning.
\newblock In {\em Proceedings of the Twenty-Fourth Conference on Artificial
  Intelligence (AAAI)}.

\bibitem[\protect\citeauthoryear{Domingos and Lowd}{2009}]{mlnbook}
Domingos, P., and Lowd, D.
\newblock 2009.
\newblock {\em Markov Logic: An Interface Layer for AI}.
\newblock San Rafael, CA: Morgan \& Claypool.

\bibitem[\protect\citeauthoryear{Fung, Mangasarian, and
  Shavlik}{2002}]{Fung02NIPS}
Fung, G.; Mangasarian, O.; and Shavlik, J.
\newblock 2002.
\newblock {Knowledge-Based support vector machine classifiers}.
\newblock In {\em NIPS},  01--09.

\bibitem[\protect\citeauthoryear{Heckerman \bgroup et al\mbox.\egroup
  }{2001}]{dn}
Heckerman, D.; Chickering, D.; Meek, C.; Rounthwaite, R.; and Kadie, C.
\newblock 2001.
\newblock Dependency networks for inference, collaborative filtering, and data
  visualization.
\newblock {\em Journal of Machine Learning Research}  49--75.

\bibitem[\protect\citeauthoryear{Kunapuli \bgroup et al\mbox.\egroup
  }{2013}]{KunapuliICDM13}
Kunapuli, G.; Odom, P.; Shavlik, J.; and Natarajan, S.
\newblock 2013.
\newblock Guiding an autonomous agent to better behaviors through human advice.
\newblock In {\em ICDM}.

\bibitem[\protect\citeauthoryear{Manning \bgroup et al\mbox.\egroup
  }{2014}]{stanfordNLP}
Manning, C.; Surdeanu, M.; Bauer, J.; Finkel, J.; Bethard, S.; and McClosky, D.
\newblock 2014.
\newblock The {Stanford} {CoreNLP} natural language processing toolkit.
\newblock In {\em Proceedings of 52nd Annual Meeting of the Association for
  Computational Linguistics: System Demonstrations},  55--60.

\bibitem[\protect\citeauthoryear{Meza-Ruiz and
  Riedel}{2009}]{SebastianRiedelJointLearning}
Meza-Ruiz, I., and Riedel, S.
\newblock 2009.
\newblock Jointly identifying predicates, arguments and senses using markov
  logic.
\newblock In {\em Proceedings of NAACL HLT}.

\bibitem[\protect\citeauthoryear{Mikolov \bgroup et al\mbox.\egroup
  }{2013}]{word2vec.iclr}
Mikolov, T.; Chen, K.; Corrado, G.; and Dean, J.
\newblock 2013.
\newblock Efficient estimation of word representations in vector space.
\newblock {\em Proceedings of Workshop at ICLR}.

\bibitem[\protect\citeauthoryear{Mikolov, Yih, and
  Zweig}{2013}]{word2vec.naacl}
Mikolov, T.; Yih, W.; and Zweig, G.
\newblock 2013.
\newblock Linguistic regularities in continuous space word representations.
\newblock {\em Proceedings of NAACL HLT}.

\bibitem[\protect\citeauthoryear{Natarajan \bgroup et al\mbox.\egroup
  }{2010}]{rdnBoost}
Natarajan, S.; Khot, T.; Kersting, K.; Gutmann, B.; and Shavlik, J.
\newblock 2010.
\newblock Boosting relational dependency networks.
\newblock In {\em Proceedings of the International Conference on Inductive
  Logic Programming (ILP)}.

\bibitem[\protect\citeauthoryear{Natarajan \bgroup et al\mbox.\egroup
  }{2014}]{ilp13WS}
Natarajan, S.; Picado, J.; Khot, T.; Kersting, K.; Re, C.; and Shavlik, J.
\newblock 2014.
\newblock Effectively creating weakly labeled training examples via approximate
  domain knowledge.
\newblock In {\em International Conference on Inductive Logic Programming}.

\bibitem[\protect\citeauthoryear{Neville and Jensen}{2007}]{rdn}
Neville, J., and Jensen, D.
\newblock 2007.
\newblock Relational dependency networks.
\newblock In {\em Introduction to Statistical Relational Learning}. The MIT
  Press.

\bibitem[\protect\citeauthoryear{Neville \bgroup et al\mbox.\egroup
  }{2003}]{rpt}
Neville, J.; Jensen, D.; Friedland, L.; and Hay, M.
\newblock 2003.
\newblock Learning relational probability trees.
\newblock In {\em In Proceedings of the ACM International Conference on
  Knowledge Discovery and Data Mining (SIGKDD)},  625--630.

\bibitem[\protect\citeauthoryear{Niu \bgroup et al\mbox.\egroup }{2011}]{tuffy}
Niu, F.; R{\'e}, C.; Doan, A.; and Shavlik, J.~W.
\newblock 2011.
\newblock Tuffy: Scaling up statistical inference in {Markov} logic networks
  using an {RDBMS}.
\newblock {\em Proceedings of Very Large Data Bases (PVLDB)} 4(6):373--384.

\bibitem[\protect\citeauthoryear{Odom \bgroup et al\mbox.\egroup
  }{2015a}]{odomAIME15}
Odom, P.; Bangera, V.; Khot, T.; Page, D.; and Natarajan, S.
\newblock 2015a.
\newblock Extracting adverse drug events from text using human advice.
\newblock In {\em Artificial Intelligence in Medicine (AIME)}.

\bibitem[\protect\citeauthoryear{Odom \bgroup et al\mbox.\egroup
  }{2015b}]{odomAAAI15}
Odom, P.; Khot, T.; Porter, R.; and Natarajan, S.
\newblock 2015b.
\newblock Knowledge-based probabilistic logic learning.
\newblock In {\em Twenty-Ninth AAAI Conference on Artificial Intelligence
  (AAAI)}.

\bibitem[\protect\citeauthoryear{Riedel, Yao, and McCallum}{2010}]{riedel10}
Riedel, S.; Yao, L.; and McCallum, A.
\newblock 2010.
\newblock Modeling relations and their mentions without labeled text.
\newblock In {\em Proceedings of the 2010 European conference on Machine
  learning and knowledge discovery in databases (ECML KDD)}.

\bibitem[\protect\citeauthoryear{Roth \bgroup et al\mbox.\egroup
  }{2014}]{RothBCGK14}
Roth, B.; Barth, T.; Chrupala, G.; Gropp, M.; and Klakow, D.
\newblock 2014.
\newblock Relationfactory: {A} fast, modular and effective system for knowledge
  base population.
\newblock In {\em Proceedings of the 14th Conference of the European Chapter of
  the Association for Computational Linguistics, {EACL} 2014, April 26-30,
  2014, Gothenburg, Sweden},  89--92.

\bibitem[\protect\citeauthoryear{Soni \bgroup et al\mbox.\egroup
  }{2016}]{akbc16}
Soni, A.; Viswanathan, D.; Pachaiyappan, N.; and Natarajan, S.
\newblock 2016.
\newblock A comparison of weak supervision methods for knowledge base
  construction.
\newblock In {\em 5th Workshop on Automated Knowledge Base Construction (AKBC)
  at NAACL}.

\bibitem[\protect\citeauthoryear{Surdeanu}{2013}]{surdeanu13}
Surdeanu, M.
\newblock 2013.
\newblock Overview of the tac 2013 knowledge base population evaluation:
  English slot filling and temporal slot filling.
\newblock In {\em Proceedings of the Sixth Text Analysis Confernece (TAC
  2013)}.

\bibitem[\protect\citeauthoryear{Towell and Shavlik}{1994}]{kbann}
Towell, G., and Shavlik, J.
\newblock 1994.
\newblock Knowledge-based artificial neural networks.
\newblock {\em Artif. Intell.} 70(1-2):119--165.

\bibitem[\protect\citeauthoryear{Zhang \bgroup et al\mbox.\egroup
  }{2012}]{zhang12}
Zhang, C.; Niu, F.; R{\'e}, C.; and Shavlik, J.
\newblock 2012.
\newblock Big data versus the crowd: Looking for relationships in all the right
  places.
\newblock In {\em Proceedings of the 50th Annual Meeting of the Association for
  Computational Linguistics: Long Papers - Volume 1}, ACL '12,  825--834.
\newblock Stroudsburg, PA, USA: Association for Computational Linguistics.

\end{thebibliography}

\end{document}